# The Use of Deep Learning in Image Segmentation, Classification and Detection

Mihai-Sorin Badea, Iulian-Ionuț Felea, Laura Maria Florea, Constantin Vertan
*The Image Processing and Analysis Lab (LAPI), Politehnica University of Bucharest, Romania*

*Abstract*—Recent years have shown that deep learned neural networks are a valuable tool in the field of computer vision. This paper addresses the use of two different kinds of network architectures, namely LeNet and Network in Network (NiN). They will be compared in terms of both performance and computational efficiency by addressing the classification and detection problems. In this paper, multiple databases will be used to test the networks. One of them contains images depicting burn wounds from pediatric cases, another one contains an extensive number of art images and other facial databases were used for facial keypoints detection.

## I. INTRODUCTION

THE increasing interest in the subject of featuredetection and classification within images has determined heightened demand regarding possible solutions in the aforementioned field. For this very purpose, challenges such as the ImageNet Large Scale Visual Recognition Challenge [1] have emerged. Challenges such as this one have shown the capabilities of deep learned networks in comparison with other classifiers (such as SVM [2]) and have participated in their elevation to mainstream status.

Another factor in the growth of deep learning is the considerable gain in computing power. The most significant boost is represented by the advent of GPU accelerated parallel computing, with platforms such as the NVIDIA CUDA [3].

For these reasons, many network architectures have been devised. Two models have been chosen to be discussed in this paper: the LeNet [4] and Network in Network (NiN)[5] models. These architectures will be used to classify data from two different databases.

The remainder of the paper is organized in the following way: first, the main aspects of the considered architectures will be presented (Section II). After this, the databases will be introduced (Section III), followed by the results obtained from the conducted experiments (Section IV). The paper is closed by a few remarks regarding the results (Section V).

## II. NETWORK ARCHITECTURE DETAILS

The first of the two proposed networks to be studied is based on the LeNet [4]Convolutional Neural Network (CNN) architecture. This model is one of the oldest ones (1998), and has been originally created for digit recognition. The general structure of a LeNet network consists in a series of a repeated blocks. Each block contains a convolutional layer followed by a pooling layer. In the used implementation each block is ended by a ReLU (Rectified Linear Unit) layer.

Following the repeated blocks, the used network contains a similar block without the pooling layer. The network is ended by a fully-connected layer and a softmax layer.

The used version of the general architecture closely resembles the Lenet-5 [4] network. In parallel, for facial keypoint detection two different networks were used: one as proposed in [6] and a modified version of it. The modified network has an extra convolutional layer that connects to each existing convolutional layer. In addition, another pair of convolutional layers have been added as a result of the parameters adjustments.

The other studied network architecture is the NiN model. As all convolutional neural networks, the NiN architecture consists in multiple blocks containing convolutional and pooling layers. The main difference between the two network types is the usage of a multilayer perceptron between the two main layers of a block. The role of the multilayer perceptron is to act as a nonlinear function approximator that can augment the network's abstraction capability [5].

A convolutional neural network can be scaled in terms of the number of blocks (automatically number of layers) or the number of feature maps during its development.

## III. THE DATABASES

As previously stated,different databases were employed for conducting the experiments, one containing images depicting burn wounds, another which contains images of paintings, and six public facial databases. Completely different in nature, all offer a satisfying number of images, considering the regular numbers of databases similar in theme.

### A. The Burn Wound Database

The images contained by this database have been obtained over the course of several months, in real hospital conditions.

A total of 611 images have been gathered from 53 pediatric patients. The age of the patients varies from 8 months to 17 years old. All the images were recorded with the same camera and originally have 1664 x 1248 pixels. All of them have been manually registered with infrared markings. The final images are 320 x 240 pixel crops of the originals, corrected with respect to the geometric distortion.

All images have been manually labeled by burn-specialized surgeons. The labels mark areas where burn wounds have occurred, along with the severity of said wounds. This way, a total of 1634 areas of interest have been detected, with severities ranging from first degree and healed burns to $3^{rd}$ degree burns.

The burn wounds by themselves were not sufficient for the experiments. As a result of this, normal skin areas have been automatically identified using the following set of rules:
1. The pixel passes a composite RGB-YUV threshold;
2. The analogous pixel in the infrared image has a corresponding temperature of over 32°C;
3. The pixel isn't already in a zone marked as a burn.

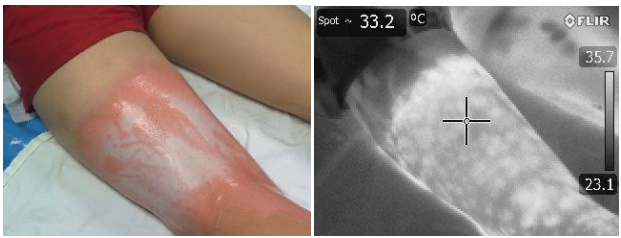

Fig. 1. Colour and IR pair from the burn wound database

### B. The paintings database

This database contains images of paintings with *Wikiart* as the main source, although more than a quarter of the database is collected from other sources. It is vast compared to other similar databases, with a total of 18040 images divided in a total of 18 art movements. All the images have been cropped to leave out the painting's framework. It was noted that some of the modern art examples contain digitized graphic.

In contrast to the burn wound database, this one does not have a constant resolution, because the paintings have different formats and their pictures were taken with different equipment.

The art movements represented in the database are:Byzantinism;Early Renaissance;Northern Renaissance;High Renaissance;Baroque;Rococo;Romanticism;Realism;Impressionism;Post-Impressionism;Expressionism;Symbolism;Fauvism;Cubism;Surrealism;Abstract art;Naïve art;Pop art.

### C. Kaggle facial keypoint detection(KFKD) database

This database contains 7049 training images, each in gray level and with a resolution of 96 x 96 pixels, and a list of 1783 test images. All the images have facial keypoints annotations, but some are missing keypoints. So only 2140 images from the training set have all the 15 keypoints annotated. The recorded images contains a large variety of illumination conditions and facial expressions.[7]

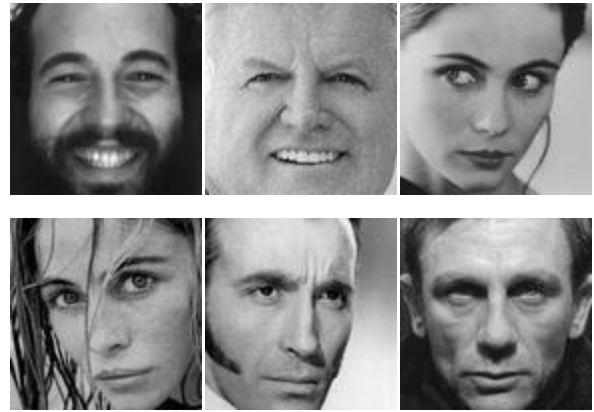

Fig. 2. Typical examples from the KFKD database

### D. BioID facial database

This database has been recorded and published to support all research in the area of face detection. It is one of the most widely used database for algorithm development and evaluation. The dataset consists of 1521 gray level images. All images havea resolution of 384 x 286 pixels. A total of 23 different subjects were used for recording the database. The recorded images are considered to be in "real world" conditions, therefore the test set features a large variety of illumination conditions, face sizes and backgrounds.[8]

All the images have been manually marked up by two PhD students, David Cristinacce and Kola Babalola. They selected 20 facial points, which are very useful for facial analysis and any further emotion detection.

### E. PUT face database

This facial database provides credible data for performance evaluation, feature extraction and recognition algorithms. It contains 2200 color images of 100 people with a resolution of 2048x1536 pixels. The images have pose variations, illumination changes, occlusions or structural disturbances. It was annotated with 65 facial points per image. [9]

### F. AR face database

This public database was created by Aleix Martinez and Robert Benavente. It contains over 4000 color images recorded from 126 subjects (70 men and 56 women). Images are of 768 x 576 pixels and of 24 bits of depth as RGB RAW files. Images depict frontal view faces with different facial expressions, illumination conditions and occlusions (sun glasses and scarves). The pictures were taken under strictly controlled conditions with no restrictions on wear, make-up, hair style, etc. The database was manually annotated with 56 facial points per image.[10]

### G. Extended Cohn-Kanade (CK+) database

This database is regularly used for research in the field of facial expression detection. In this purposes, it is one of the most popular databases to date. It contains over 11000 images with a resolution of 640x490 or 640x480 pixels in grayscale. It contains 593 sequences from 123 subjects. Participants were 18 to 50 years of age, 69% female. Each participant was instructed to perform a series of 23 facial displays, which included single action units and combinations of action units. The image sequences incorporate images form onset (neutral frame) to peak expression (last frame). The database was manually annotated with 68 facial points per image. [11]

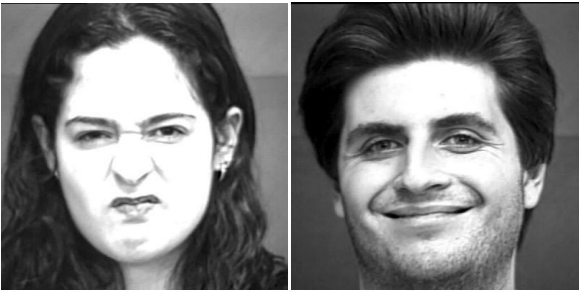

Fig. 3. Examples from the CK+ database depicting emotions on apex

### H. Labeled Faces in the Wild (LFW) database

The LFW database contains 13233 facial images of 5749 individuals, 1680 of which have more than one image in the database. The images have been collected "in the wild" and vary in pose, illumination, resolution, quality, expression, gender, age, race, occlusion and make-up. All the images are colored and have a resolution of 250 x 250 pixels.[12]

All faces from LFW database have 10 facial feature points annotated by Amazon Mechanical Turk.

## IV. EXPERIMENTS

Using the databases described above, a series of experiments were conducted to study the effectiveness of CNN in various conditions and implementations. The results were compared with those obtained through the use of other classifiers. Along with experiments regarding the pure classification/detection performance of the networks, another was devised to measure the speed of training.

As far as the performance tests go, the networks were trained to discern between regular and burned skin, or between regular skin, light burns and serious burns, in the case of the burn wound database. The first case uses only RGB information, while the latter takes the infrared information into account. All experiments on the burn wound database use small patches extracted from the images. Thus, the global classification problem is transformed into a local, small-scale one. In this way, the CNNs were used for a segmentation problem as well.

When considering the paintings database, the networks were trained to differentiate between different art movements. Taking into account that the database contains images of different resolutions and formats, all samples had to be resized to accommodate the inputs of different networks.

For facial points detection the networks were trained to detect location of facial keypoints. The networks use as a training and validation set images from the KFKD database and an augmented version of it. As for the test set, images from all other facial databases were used.

The experiments were conducted using Matlab along with the MatConvNet [13] toolbox when considering the first two databases. The tested architectures are based on the example networks provided by the toolbox. The Lasagne and Theano libraries were used with Python for the facial databases.

### A. Influence of networks architectures and input size on classification performance

Since the LeNet-5 architecture was used for OCR on 32 x 32 images [4] and the original NiN architecture was used on 32 x 32 databases (CIFAR-10, CIFAR-100, SVHN) [5], both architectures were trained using 32 x 32 patches extracted from the images contained by the burn wound database. All patches extracted from the burn wound database have been selected as to contain pixels of interest (normal or burned skin).

The networks trained for the paintings database, on the other hand, don't use patches. They used scaled versions of the paintings. After this initial test, some of the networks were modified to accommodate an input of 64 x 64 pixels. It is mentioned that the NiN network was not used in the 64 x 64 format on the burn wound database, or the 32 x 32 format on the paintings database. As for the LeNet network, it was not used in its 64 x 64 format in the regular skin vs. burned skin classification case.

Moving to the detection problem, the input size was not modified. The networks differ by the number of convolutional layers, with adjusted parameters. Thus, the effect of the depth of the networks was studied, in detriment to the input size. In this case, performance is considered to be the percentage of images that have a prediction error for the facial keypoints lower than 0.1. The prediction error is measured as a fraction of the inter-ocular distance. The results can be seen in Table I.

It can be seen that the NiN networks are slightly better than the LeNet ones. That was to be expected considering that NiN networks are basically improved LeNet architectures. It is worth mentioning that the NiN networks tested had almost the same results when using dropout layers in comparison to not using them.

Table I
PERFORMANCE COMPARISON

| Problem | Database | Network | Input Size | Performance [%] |
|---|---|---|---|---|
| Skin vs Burn | Burn database | LeNet | 32 x 32 | 75.91 |
| Skin vs Light Burn vs Serious Burn | Burn database | LeNet | 32 x 32 | 58.01 |
| | | LeNet | 64 x 64 | 53.68 |
| | | NiN | 32 x 32 | 55.7 |
| Art Movement | Paintings database | LeNet | 32 x 32 | 22.3 |
| | | LeNet | 64 x 64 | 25.1 |
| | | NiN | 64 x 64 | 26.5 |
| Facial Keypoint Detection | BioID | LeNet | 96 x 96 | 99.79 |
| | | Modified LeNet | 96 x 96 | 100 |
| | PUT | LeNet | 96 x 96 | 98.20 |
| | | Modified LeNet | 96 x 96 | 99.43 |
| | AR | LeNet | 96 x 96 | 99.6 |
| | | Modified LeNet | 96 x 96 | 100 |
| | CK+ | LeNet | 96 x 96 | 2.89 |
| | | Modified LeNet | 96 x 96 | 37.77 |
| | LFW | LeNet | 96 x 96 | 71.61 |
| | | Modified LeNet | 96 x 96 | 76.57 |

The more interesting fact is represented by the performance of the 64 x 64 version of the networks. Even though both architectures scored better in 64 x 64 versions in comparison to the LeNet 32 x 32 version, in the skin vs light burn vs serious burn context this was not the case.

The most probable reason behind the lower score in the 64 x 64 LeNet case is the nature of the picture. All types of zones have high spatial redundancy in this database. Extending the patch to 64 x 64 pixels makes it more likely to mix areas of interest with clutter from the background, making the classification problem harder. Moreover, many burn areas are relatively small and a larger patch could lead to a smaller proportion of relevant information in the patch.

The skin vs burn case has good results, but is the easiest of all three classification problems.

The facial keypoint detection problem has underlined the effects of a larger CNN. The additional layers and adjustments have brought a visible boost in performance on all considered databases. In some cases, even the unmodified LeNet architecture has extremely good results. The difference between the two LeNet models was clear when considering extreme facial expressions or ones observed "in the wild", such as the ones in the CK+ and LFW databases. These results are heavily influenced by the fact that no training image contais such facial expressions.

It is worth mentioning that results in this case are obtained in real-time.

*B. Comparison to other classifiers*

The results presented in the previous subsection must be compared to those obtained from other classifiers for a better frame of reference. The regular skin vs light burn vs serious burn case has not been compared to other classifiers.

The skin vs burn results have been compared to a classification using a Hue-Saturation model [14] and a Color-Texture model [15].

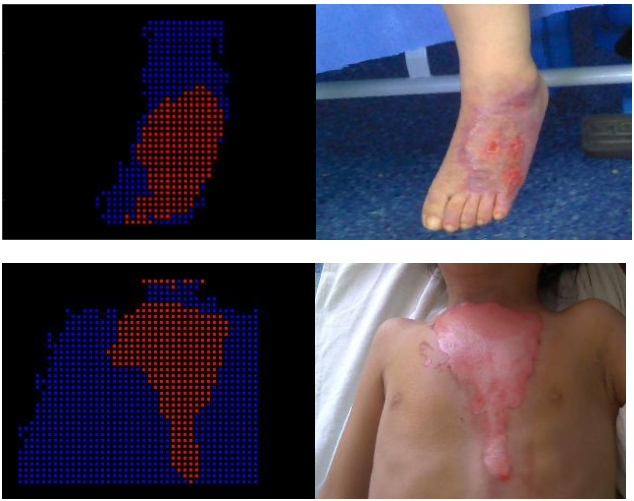

Fig. 4. Examples of well classified images in the skin vs burn case Blue blocks represent skin, while the red ones represent burns.

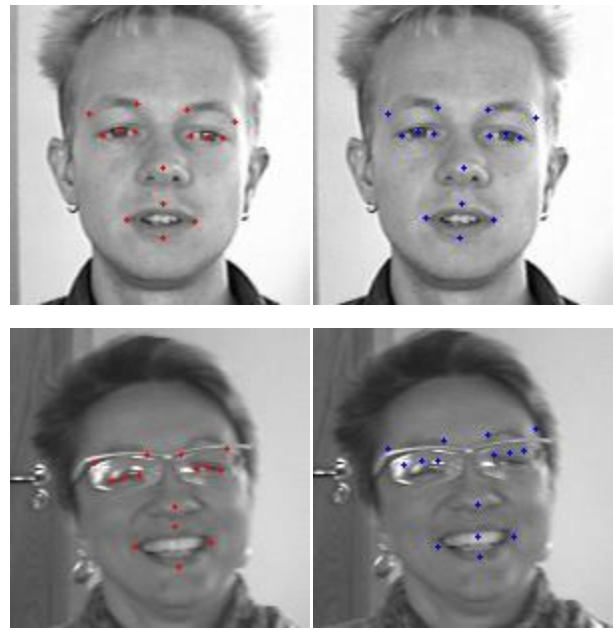

Fig. 5. Cropped examples of well detected (top) and badly detected keypoints (bottom) using the unmodified LeNet architecture.. The left side shows the ground truth and the right side shows the predicted keypoints.

The art movement classification using CNN has been compared with a ResNet network (with a 224 x 224 input size) [16], a SVM classifier and a Random Forest classifier. The last two use the following feature extractors: HoT [17], HoG, LPB, LIOP, EHD, HTD, SIFT.

The results obtained from the facial keypoint detection problem have been compared considering the BioID database. The comparison was made with the BoRMaN method proposed in [18] and the method proposed by Belhumeur et al. in [19]. Table II contains the comparison between all the classifiers.

It can be easily seen that, in the skin vs burn experiment, the tested LeNet network has outperformed the other classifiers. Again, it is the easiest problem, and other, more complex classifiers such as the ones used for the next case aren't fit for pixel by pixel classification.

Table III
PERFORMANCE COMPARISON

| Classification problem | Method | Performance [%] |
|---|---|---|
| Skin vs Burn | LeNet | 75.91 |
| | Hue-Saturation | 53.87 |
| | Color-Texture | 65.34 |
| Art Movement | NiN | 26.5 |
| | ResNet | 47.8 |
| | RF | 44.5 |
| | SVM | 50.0 |
| Facial Keypoint Detection (BioID) | Modified LeNet | 100 |
| | BoRMaN [18] | ≈ 95 |
| | Belhumeur [19] | ≈ 97 |

On the other hand, in the painting classification problem, the tested NiN falls behind all other tested classifiers. This is probably caused by the nature of the problem. The picture needs to be classified as a whole, and such a small patch isn't fit to contain enough visual information from the image. All other classifiers have access to a unmodified version of the image. The Random Forest and SVM classifiers are advantaged by the global characteristics.

When considering the facial keypoint detection problem, the modified LeNet architecture's performance isn't matched by other considered methods on the BioID database.

*C. Speed Comparison*

An important aspect of CNN is represented by the speed of training. Large data sets lead to extremely large training times. Besides this aspect, a small evaluation time could mean feasibility in real time applications. As stated in the opening an important factor in the evolution of deep learning is represented by the present technologic capabilities. GPU accelerated parallel computing is a valuable tool in the development of a deep learned neural network.

All the training has been conducted on two personal computers with the following specifications:Computer 1:CPU: Intel Core i5 760 @2.8GHz;RAM: 12 GB;GPU: NVIDIA GeForce GTX 970, 4096 MB.Computer 2:CPU: Intel Xeon E5-1620 @3.6GHz;RAM: 24GB;GPU: NVIDIA GeForce GTX 980 ti, 4096 MB;

The MatConvNet toolbox allows the use of GPU accelerations. Thus, the speedup gained through the use of the video card has been measured by training the same network on both computers, with GPU acceleration on and off. For this, a small database consisting of 32 x 32 pixel patches has been created, as to keep training times low.Fig. 6 shows the relative speedup reported to the longest running time.

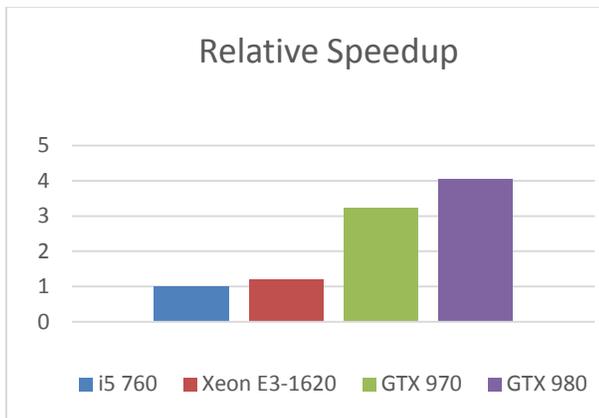

Fig. 6. The relative speedup when using different video cards and CPUs. The i5 760 was used as a reference.

An incredible difference can be seen between the CPU-only versions and the ones that employed GPU acceleration. A relevant difference can be seen between the two video cards as well, with an almost 26% boost for the GTX 980. Bigger databases and network architectures will only make the gap more visible between the CPU-only and GPU accelerated versions.

V. CONCLUSIONS

The conducted experiments have shown the differences between different architectures and the impact of the GPU on the development of deep learned networks.Simple architectures such as LeNet and NiN have proven themselves reliable for low complexity classification but not sufficient for more difficult tasks. This does not mean that deep learned networks are not adequate, judging by the more than satisfactory results of ResNet.

When considering the task of facial keypoint detection, the LeNet architecture has satisfying results in most cases but is always surpassed by the modified version. The larger input size of a network may not guarantee better results as seen in the second classfication problem considered on the burn wound database. Still, a larger input size allows the addition of extra layers, which can lead to better results.